# Proliferating cell nuclear antigen (PCNA) allows the automatic identification of follicles in microscopic images of human ovarian tissue

Thomas W Kelsey[1]
Benedicta Caserta[2]
Luis Castillo[2]
W Hamish B Wallace[3]
Francisco Cóppola Gonzálvez[4]

[1]School of Computer Science, University of St Andrews, St Andrews, Scotland, UK; [2]Pereira Rossell Hospital, ASSE, Ministry of Public Health, Montevideo, Uruguay; [3]Division of Child Life and Health, Department of Reproductive and Developmental Sciences, University of Edinburgh, Edinburgh, Scotland, UK; [4]Department of Obstetrics and Gynecology C, School of Medicine, University of the Republic, Montevideo, Uruguay

Correspondence: Tom Kelsey
School of Computer Science, University of St Andrews, North Haugh, St Andrews, KY16 9SX United Kingdom
Tel +44 1334 463249
Fax +44 1334 463278
Email tom@cs.st-andrews.ac.uk

**Background:** Human ovarian reserve is defined by the population of nongrowing follicles (NGFs) in the ovary. Direct estimation of ovarian reserve involves the identification of NGFs in prepared ovarian tissue. Previous studies involving human tissue have used hematoxylin and eosin (HE) stain, with NGF populations estimated by human examination either of tissue under a microscope, or of images taken of this tissue.
**Methods:** In this study we replaced HE with proliferating cell nuclear antigen (PCNA), and automated the identification and enumeration of NGFs that appear in the resulting microscopic images. We compared the automated estimates to those obtained by human experts, with the "gold standard" taken to be the average of the conservative and liberal estimates by three human experts.
**Results:** The automated estimates were within 10% of the "gold standard", for images at both 100× and 200× magnifications. Automated analysis took longer than human analysis for several hundred images, not allowing for breaks from analysis needed by humans.
**Conclusion:** Our results both replicate and improve on those of previous studies involving rodent ovaries, and demonstrate the viability of large-scale studies of human ovarian reserve using a combination of immunohistochemistry and computational image analysis techniques.

## Introduction

The human ovary contains a fixed number of nongrowing follicles (NGF) established before birth. This number declines with increasing age culminating in the menopause at 50–51 years.[1] Ovarian reserve is defined by the remaining population at a given age. There is no technique known for direct *in vivo* estimation of ovarian reserve; indirect indicators include antral follicle counts, ovarian volume and levels of hormones such as follicle-stimulating hormone and anti-Mullerian hormone.[2] A model describing the age-related population of NGFs in the human ovary from conception to menopause, in which the NGF populations of the 325 ovaries studied were all estimated using variations on the standard methodology developed by Block in the early 1950s,[4,5] has recently been published.[3] After oophorectomy (or post-mortem) the ovary is fixed, thin slices (between 5 and 20 microns) are taken at regular intervals, and these are stained with hematoxylin and eosin (HE). Sample regions are either inspected manually, or photographed, with the NGFs appearing in the tissue counted by hand.

**Table 1** Sample comparison of automated counts to human counts

| Mag. | Image | Human 1 | | | Human 2 | | | Human 3 | | | Automated | | | Human Mean |
|---|---|---|---|---|---|---|---|---|---|---|---|---|---|---|
| | | Con | Lib | Mean | Con | Lib | Mean | Con | Lib | Mean | Con | Lib | Mean | |
| 200 | 1 | 0 | 0 | 0 | 0 | 0 | 0 | 0 | 0 | 0 | 0 | 0 | 0 | 0.0 |
| | 2 | 0 | 0 | 0 | 0 | 0 | 0 | 0 | 0 | 0 | 0 | 0 | 0 | 0.0 |
| | 3 | 0 | 0 | 0 | 0 | 0 | 0 | 0 | 0 | 0 | 0 | 1 | 0.5 | 0.0 |
| | 4 | 0 | 0 | 0 | 0 | 0 | 0 | 0 | 0 | 0 | 0 | 0 | 0 | 0.0 |
| | 5 | 0 | 0 | 0 | 0 | 0 | 0 | 0 | 0 | 0 | 0 | 0 | 0 | 0.0 |
| | 6 | 0 | 1 | 0.5 | 0 | 1 | 0.5 | 0 | 1 | 0.5 | 0 | 1 | 0.5 | 0.5 |
| | 7 | 0 | 2 | 1 | 0 | 2 | 1 | 0 | 2 | 1 | 0 | 2 | 1 | 1.0 |
| | 8 | 0 | 2 | 1 | 0 | 2 | 1 | 0 | 3 | 1.5 | 0 | 2 | 1 | 1.2 |
| | 9 | 0 | 3 | 1.5 | 0 | 3 | 1.5 | 0 | 3 | 1.5 | 0 | 3 | 1.5 | 1.5 |
| | 10 | 6 | 8 | 7 | 5 | 6 | 5.5 | 6 | 7 | 6.5 | 5 | 9 | 7 | 6.3 |
| | 11 | 0 | 0 | 0 | 0 | 0 | 0 | 0 | 0 | 0 | 0 | 0 | 0 | 0.0 |
| | 12 | 0 | 2 | 1 | 0 | 0 | 0 | 0 | 0 | 0 | 0 | 2 | 1 | 0.3 |
| | 13 | 0 | 0 | 0 | 0 | 0 | 0 | 0 | 0 | 0 | 0 | 0 | 0 | 0.0 |
| | 14 | 0 | 0 | 0 | 0 | 0 | 0 | 0 | 0 | 0 | 0 | 1 | 0.5 | 0.0 |
| | 15 | 5 | 7 | 6 | 0 | 6 | 3 | 3 | 6 | 4.5 | 0 | 6 | 3 | 4.5 |
| | 16 | 2 | 5 | 3.5 | 1 | 3 | 2 | 2 | 3 | 2.5 | 1 | 5 | 3 | 2.7 |
| | 17 | 1 | 1 | 1 | 0 | 1 | 0.5 | 0 | 1 | 0.5 | 0 | 1 | 0.5 | 0.7 |
| | 18 | 0 | 0 | 0 | 0 | 0 | 0 | 0 | 0 | 0 | 0 | 0 | 0 | 0.0 |
| | 19 | 0 | 0 | 0 | 0 | 0 | 0 | 0 | 0 | 0 | 0 | 2 | 1 | 0.0 |
| | 20 | 0 | 0 | 0 | 0 | 0 | 0 | 0 | 0 | 0 | 0 | 0 | 0 | 0.0 |
| | 21 | 0 | 0 | 0 | 0 | 0 | 0 | 0 | 0 | 0 | 0 | 0 | 0 | 0.0 |
| | 22 | 1 | 8 | 4.5 | 1 | 6 | 3.5 | 2 | 6 | 4 | 1 | 8 | 4.5 | 4.0 |
| | 23 | 4 | 5 | 4.5 | 3 | 5 | 4 | 3 | 5 | 4 | 3 | 5 | 4 | 4.2 |
| | 24 | 2 | 4 | 3 | 2 | 2 | 2 | 2 | 3 | 2.5 | 2 | 4 | 3 | 2.5 |
| | 25 | 2 | 6 | 4 | 2 | 3 | 2.5 | 2 | 5 | 3.5 | 2 | 6 | 4 | 3.3 |
| | 26 | 1 | 3 | 2 | 0 | 1 | 0.5 | 1 | 3 | 2 | 0 | 3 | 1.5 | 1.5 |
| | 27 | 0 | 3 | 1.5 | 0 | 0 | 0 | 0 | 2 | 1 | 0 | 3 | 1.5 | 0.8 |
| | 28 | 1 | 3 | 2 | 1 | 2 | 1.5 | 1 | 3 | 2 | 1 | 3 | 2 | 1.8 |
| | 29 | 1 | 1 | 1 | 1 | 1 | 1 | 1 | 1 | 1 | 1 | 1 | 1 | 1.0 |
| | 30 | 0 | 8 | 4 | 0 | 3 | 1.5 | 0 | 6 | 3 | 0 | 11 | 5.5 | 2.8 |
| | 31 | 5 | 7 | 6 | 4 | 6 | 5 | 5 | 6 | 5.5 | 4 | 7 | 5.5 | 5.5 |
| | 32 | 0 | 0 | 0 | 0 | 0 | 0 | 0 | 0 | 0 | 0 | 0 | 0 | 0.0 |
| | 33 | 0 | 4 | 2 | 0 | 2 | 1 | 0 | 3 | 1.5 | 0 | 4 | 2 | 1.5 |
| | 34 | 1 | 1 | 1 | 0 | 1 | 0.5 | 1 | 1 | 1 | 0 | 1 | 0.5 | 0.8 |
| | 35 | 0 | 3 | 1.5 | 0 | 2 | 1 | 0 | 3 | 1.5 | 0 | 3 | 1.5 | 1.3 |
| | 36 | 0 | 0 | 0 | 0 | 1 | 0.5 | 0 | 0 | 0 | 0 | 0 | 0 | 0.2 |
| | 37 | 0 | 2 | 1 | 0 | 1 | 0.5 | 0 | 2 | 1 | 0 | 2 | 1 | 0.8 |
| | 38 | 4 | 4 | 4 | 2 | 5 | 3.5 | 4 | 5 | 4.5 | 2 | 4 | 3 | 4.0 |
| | 39 | 1 | 2 | 1.5 | 0 | 1 | 0.5 | 1 | 1 | 1 | 0 | 2 | 1 | 1.0 |
| | 40 | 1 | 1 | 1 | 0 | 1 | 0.5 | 0 | 1 | 0.5 | 0 | 1 | 0.5 | 0.7 |
| | 41 | 9 | 13 | 11 | 7 | 11 | 9 | 7 | 13 | 10 | 6 | 11 | 8.5 | 10.0 |
| | 42 | 0 | 0 | 0 | 0 | 0 | 0 | 0 | 0 | 0 | 0 | 0 | 0 | 0.0 |
| | 43 | 2 | 5 | 3.5 | 0 | 2 | 1 | 1 | 3 | 2 | 0 | 5 | 2.5 | 2.2 |
| | Total | 49 | 114 | 81.5 | 29 | 80 | 54.5 | 42 | 98 | 70 | 28 | 119 | 73.5 | 68.7 |

**Notes:** Three human experts performed both liberal and conservative NGF population estimates from 42 microscopy images of a single human ovary taken at 200× magnification. The average of the averages of these counts are given in the final column; the average of the automated estimates are given in the penultimate column.
**Abbreviations:** Mag, magnification; Con, conservative estimate; Lib, liberal estimate.

Assuming an even distribution of NGFs throughout the ovary, the full population is then estimated using solutions of the corpuscle problem for 3-dimensional specimens. This process is time consuming, and suffers from human misclassification, integration error due to small sample sizes, and the inconsistent assumption of even distribution. In his seminal paper from 1952,[4] Block provided the motivation for this study:

*The distribution of these follicles in human ovaries is so uneven that reliable values cannot be obtained until all the*

**Table 2** Summarized comparison of automated counts to human counts for microscopy images

| Mag. | No. Images | Human 1 | | | Human 2 | | | Human 3 | | | Automated | | | Human Mean |
|---|---|---|---|---|---|---|---|---|---|---|---|---|---|---|
| | | Con | Lib | Mean | Con | Lib | Mean | Con | Lib | Mean | Con | Lib | Mean | |
| 100 | 220 | 191 | 399 | 295 | 182 | 377 | 279.5 | 180 | 370 | 275 | 189 | 416 | 302.5 | 283.2 |
| 200 | 97 | 70 | 211 | 140.5 | 69 | 201 | 135 | 83 | 206 | 144.5 | 73 | 230 | 151.5 | 140.0 |

**Notes:** Three human experts performed both liberal and conservative NGF population estimates from microscopy images of PCNA stained sections from three human ovaries at 200× and 100× magnifications. The average of the averages of these counts are given in the final column; the average of the automated estimates are given in the penultimate column.
**Abbreviations:** Mag, magnification; No, number of; Con, conservative estimate; Lib, liberal estimate.

*follicles are counted. This requires complete serial sectioning, which for a woman of fertile age means one thousand five hundred to two thousand five hundred 20 micron sections per ovary. Under these circumstances any large-scale investigation is impracticable....*

Our aim is to automate part of the estimation process, so that the use of modern image preparation and analysis tools and techniques can reduce the human workload involved in more accurate ovarian reserve studies. We report a combined process of tissue staining and automatic feature detection, which gives results comparable to human counts. Our process works at low magnifications (thereby reducing the number of images needed per section), and can, in principle, be used to obtain almost exact NGF populations from fully sectioned ovaries.

## Material and methods

We studied biopsies of tissue from three intact ovaries (post-oophorectomy) serially sectioned, obtained after routine surgery for cancer patients. None of the subjects had cancer of the ovary. The ages of the patients for whom oophorectomy was performed were 12, 18, and 20 years. Ovarian tissue was received unfixed from theatre, and was on the same day fixed in buffered formalin for between 24 and 48 hours and embedded in paraffin. At a later date, the ovaries were sectioned into 10–12 slices, from which 5 microns thick slide tissue was obtained using a Microtomo knife (Leitz GmbH and Co KG, Baden-Wurtemberg, Germany).

## Immunohistochemistry

Proliferating Cell Nuclear Antigen (PCNA) was used as the primary stain, in line with a successful study on rodent ovaries.[6] Our tissue preparation methods differ from this study only in that we counterstained with hematoxylin for 60 seconds rather than three minutes and we used 1:100 dilution of PCNA (instead of 1:400) as recommended by the stain supplier (BioCare Medical LLC, California, USA). The preparation sequence was

1. Dis-paraffination and hydration
2. Heat induced Antigen retrieval for 60 minutes with tris(hydroxymethyl)aminomethane-ethylenediaminet-etraacetic acid (4 molar TRIS-EDTA) buffer solution (pH 9)
3. Wash in distilled water 10 minutes followed by buffer wash (0.1 molar phosphate buffered saline (PBS))
4. Incubation at room temperature for 60 minutes with primary antibody (mouse monoclonal PCNA concentrate, dilution 1:100, clone PC 10 (BIOCARE)) followed by buffer wash (PBS)
5. Incubation using Dual Link Heat-Stable Protein (HPr) (DAKO Envision™ (DAKO Denmark, Glostrup, Denmark)) for 30 minutes at room temperature followed by buffer wash (PBS)
6. Application of diaminobenzidine (DAB) chromogen (DAKO Denmark A/S) for 10 minutes followed by wash in distilled water
7. Counterstain nuclear-Mayer hematoxylin for 1 minute followed by 10 minutes under running water
8. Dehydration in alcohol
9. Application of Xylene
10. Mounting on a standard coverslipped slide

External positive controls were performed on tissue from patients with breast cancer and cancer of the colon.

## Image preparation

Slides were viewed using a 1.3 megapixel Infinity 1 camera (Lumanera Corp., Ottawa, Canada) attached to an Olympus CX31 microscope (Olympus Imaging Corp., Tokyo, Japan). Images were captured using the Infinity Analyze software package supplied with the camera. The default exposure, hue, saturation, brightness and contrast settings were used. The white balance was adjusted from a default of Red 2.28, Green 1.80, Blue 2.69 to Red 1.84, Green 1.84, Blue 3.30. Light intensity was set to 3.6 for 100× images, and to 4.0 for 200× images. Images were saved as 32-bit-per-pixel RGB 1280 × 1024 pixel TIFF files.

## Image analysis

Our computational methodology varied with the magnification used. We used the ImageJ (National Institutes of Health, Washington DC, USA) suite of image analysis tools throughout, making extensive use of the morphology software extensions developed by Dr G. Landini of the University of Birmingham, Birmingham, UK.

For 200× images we: (1) compute the maximum entropy threshold, (2a) identify regions of restricted size having low aspect ratio, high modulus ratio, high sphericity, and which do not contain too much blue, (2b) identify regions that have some circularity, for which blue average is not low, and which are not background (ie, green and red kurtosis are positive), (3) combine the two sets of isolated regions. Processes 2a and 2b isolate NGF nuclei and zona pellucida (ZP) respectively. Parts of the image that consist of nucleus plus ZP are classified as NGFs, as are regions consisting only of ZP: these are NGFs that have been sliced in an area not containing the nucleus. Isolated nuclei thus represent false positives and are discarded.

For 100× images we: (1) compute the triangle entropy, (2) identify regions of restricted size as in 2a above, (3) filter out any particles with low compactness and circularity and/or high aspect ratio (values chosen are liberal, since we apply a color filter to the survivors), and (4) filter by color: median RGB must be lower than 70, 60, and 55, respectively (giving a very dark brown). For both magnifications, we run the code twice – with liberal and conservative settings – and take the average as our estimate of the number of NGFs in the image.

## NGF counts by hand

Laboratory staff performed two counts for each image. One conservative (including only those regions of the image that certainly represented NGFs), and the other liberal (including both definite NGFs and regions that could equally be either an NGF or a sectioned blood/lymphatic vessel). These counts were added together, and the average taken.

## Results

We obtained excellent results for both 200× (Figure 1) and 100× images (Figures 2 and 3). For this small sample, the automatic identification code with conservative settings consistently agreed to within 5% with the average conservative human count. With liberal settings the code consistently agreed to within 10%. Taking the average of these counts (both human and automatic) to be a good estimate of the true number of NGFs present, the automated image analysis count was indistinguishable from averages of expert human counts, being neither more conservative nor more liberal than the average human counter. There was wider variance in population estimates at 100× for both human and automated image analysis counters.

The automated analysis was, on average, a factor of two times slower than the time taken by human experts. However, these timings do not take into account the breaks needed for a human when analyzing tens of thousands of images. If we assume that a human can work accurately for less than 12 hours per day, then the automated analysis becomes the faster method.

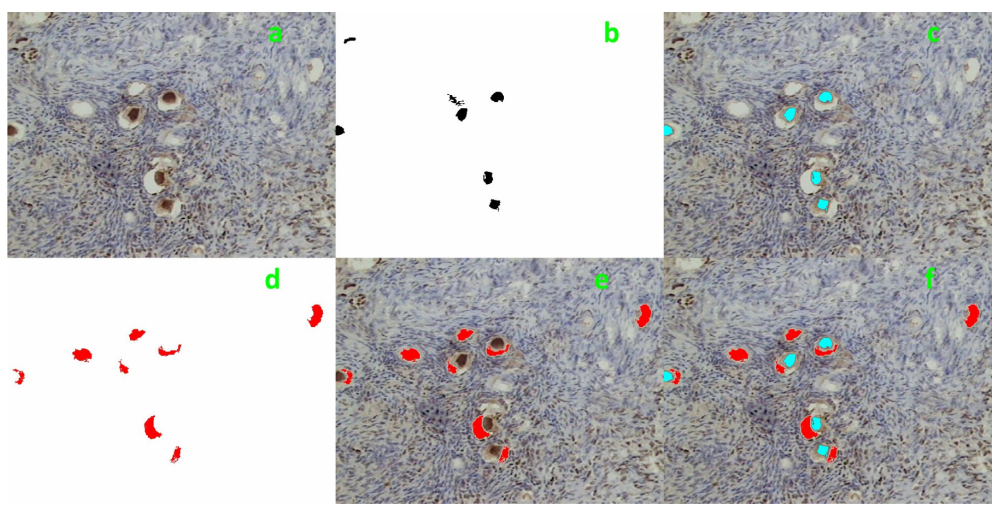

**Figure 1** Automatic NGF identification in PCNA stained human ovarian tissue (original image taken at 200× magnification) with liberal settings. Panel (a) is the original image. Panels (b) and (c) show the identification of NGF nuclei by color, size and shape. Panels (d) and (e) show the identification of light areas (either ZP or sectioned blood/lymphatic vessels), also by color, size and shape. Panel (f) shows the identified NGFs with liberal settings applied: a light area of the correct size and shape is classified as an NGF that has not been sectioned through the nucleus. Human expert estimates for the number of NGFs in this image range from 5 to 8.

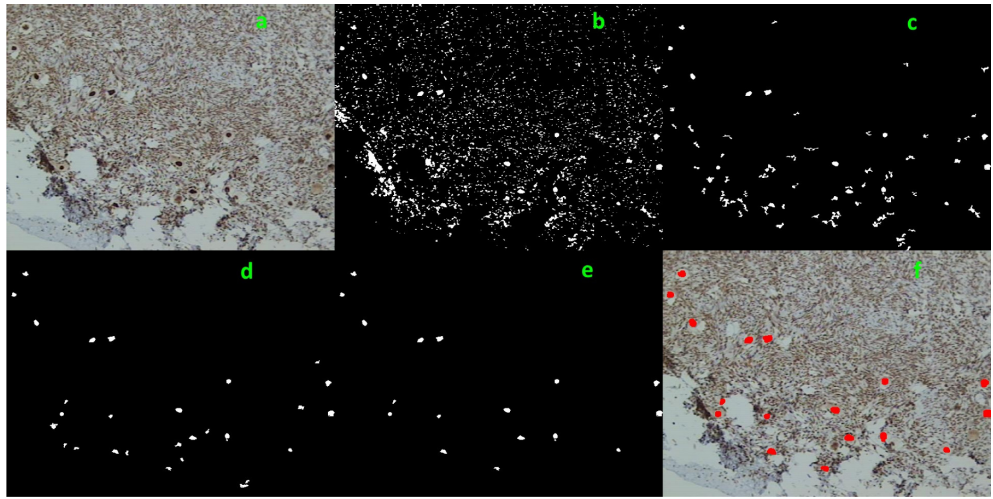

**Figure 2** Automatic NGF identification in PCNA stained human ovarian tissue (original image taken at 100× magnification) with liberal settings. Panel (a) is the original image. Panel (b) shows the result of triangle thresholding. Panels (c) through (e) show filtering by size, shape and color respectively. Panel (f) shows 17 identified NGFs with liberal settings applied. Human expert estimates for the number of NGFs in this image range from 14 to 17.

## Discussion

Ovarian tissue consists of stroma cells, NGFs – consisting of an oocyte surrounded by zona pellucida (ZP) – and growing follicles, supported by an extracellular matrix.[7] Blood and lymphatic vessels are also present. The standard stain, HE, hinders computational image analysis even at high magnification, since sub-regions of NGFs can have the same color (and size and morphology) as stroma cells held in the extracellular matrix. Moreover, an obvious candidate as a computational technique – color deconvolution into shades of pink and blue – cannot be fully automated since HE is a nonstoichiometric stain, and hence *a priori* empirical derivation of stain vectors is needed for (at least) each batch of images.

PCNA, however, stains the nuclei of the stroma cells and NGFs in shades of brown since these cells are in the $G_1$, S or $G_2$ interphase stages of cell development. The nuclei of NGFs are typically stained a darker brown than the nuclei of the stroma cells, allowing us to differentiate by color as well as morphology and size. The slight hematoxylin counterstain that we used gives a blue color to the extracellular matrix, leaving the ZP an almost unstained light color. This distinction of regions of images by color allows us to add color differentiation to the size and morphology

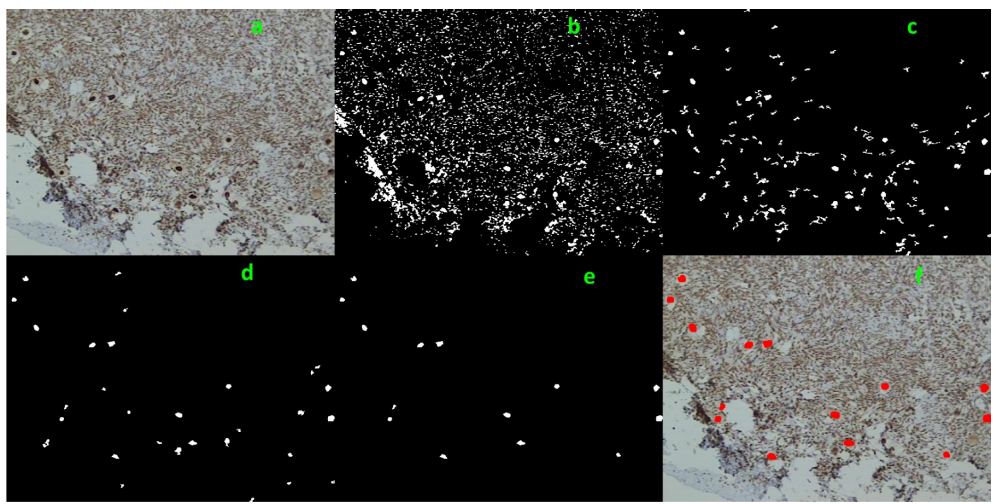

**Figure 3** Automatic NGF identification in PCNA stained human ovarian tissue (original image taken at 100× magnification) with conservative settings. Panel (a) is the original image. Panel (b) shows the result of triangle thresholding. Panels (c) through (e) show filtering by size, shape, and color, respectively. Panel (f) shows 14 identified NGFs with conservative settings applied. Human expert estimates for the number of NGFs in this image range from 14 to 17.

attributes that are the only tools available for HE-stained tissue.

Human identification of NGFs is far from unambiguous. A study involving five monkey ovaries[8] reported average populations of 15,735 NGFs, with a standard deviation of 6,214. A study involving 10 rat ovaries[6] reported average NGF populations of 871 (SD, 279) with HE stain, and 1132 (SD, 290) using PCNA. Both studies reported a normal distribution of estimates, indicating no human tendency to consistently over- or underestimate the true population. The problem is therefore the precision of the estimates rather than their accuracy, and hence averaging multiple counts is almost certain to be more accurate than a single count. However, the inherent uncertainty in individual estimates hinders the reporting of exact results when comparing human counts with those obtained by a computer program. Our approach in this study, therefore, was to mimic two human observers (one conservative, the other liberal) and estimate the true population as the average of these two counts. It should be noted, however, that users of the code who prefer to count, for example, only textbook examples of NGFs may simply use the conservative settings and take the results of these counts as their population estimate.

Previous studies have investigated the use of computational techniques to estimate NGF populations in images of rodent ovarian tissue. The study on rats[6] published in 2008 provides no details on the image analysis performed, but does refer to it as semiautomatic rather than automatic. A more recent study[8] involving mouse ovaries stained with mouse vasa homolog (MVH) generates comparable data with conventional methods of NGF counting, and the authors provide a full description of the imaging techniques used. This study reports a semiautomatic rather than automatic image analysis method, noting that light micrographs will differ from dark micrographs, so that computer settings have to be altered for each batch of images depending on the intensity of the staining of the nuclei in that batch. The results obtained for our study were completely automatic: no human input was required to adjust settings before the automated counts (Tables 1 and 2). This could, however, be due to the small number of ovaries that we examined, and it may well be the case that, in general, a level of human involvement is needed to pre-process a batch of images before accurate automatic counting can proceed.

The main strength of our study is that we have used human ovarian tissue. The main limitation of this study is the small number of slides examined, from a small number of ovaries. Clearly, rodent ovaries are more easily obtainable than those of human subjects, but our aim is to address the more important research question of how best to estimate human ovarian reserve.

## Conclusions

To our knowledge, we present the first combination of PCNA staining combined with fully automated image analysis to estimate human NGF populations from histological images. Neither of our methods (for images taken at 200× and 100×) requires pre-processing before use: the thresholding automatically gives good results for our PCNA stained tissue. By running our code on a cluster of computational nodes, it is entirely feasible to automatically estimate NGF populations from all the images obtained from every section of a human ovary. It may be possible that differences in stain levels across many ovaries, and/or across multiple laboratories will mean that some human input is needed to regularize each batch of images, as found by the recent study involving mice.[9] Further research is needed into this possibility.

## Acknowledgments/disclosures

The authors report no conflicts of interest in this work. TWK is supported by EPSRC grants EP/CS23229/1 and EP/H004092/1. The funders had no role in study design, data collection and analysis, decision to publish, or preparation of the manuscript. We would like to acknowledge the critical discussions with, and the technical support from, A Sica, M Cedeira, D Mazal, S de la Peña, S Bianco, V Carbonati, C Carrera, M Correa, R Gabrielli, R Nesti and M Cuello. We thank Mr TM Shariah Sazzad and Ms P Wright for help with the human counting of NGFs.